%% file: main.tex
\documentclass[10pt,twocolumn,letterpaper]{article}

\usepackage{cvpr}              
\usepackage{multirow}

\input{preamble}

\definecolor{cvprblue}{rgb}{0.21,0.49,0.74}
\usepackage[pagebackref,breaklinks,colorlinks,citecolor=cvprblue]{hyperref}

\def\paperID{*****} 
\def\confName{CVPR}
\def\confYear{2024}

\title{Enhancing Traffic Safety with Parallel Dense Video Captioning for End-to-End Event Analysis}

\author{Maged Shoman\\
Smart and Safe Transportation (SST) Lab\\
Department of Civil, Environmental and Construction Engineering\\
University of Central Florida, United States\\
{\tt\small Maged.shoman@ucf.edu}
\and
Dongdong Wang\\
Smart and Safe Transportation (SST) Lab\\
Department of Civil, Environmental and Construction Engineering\\
University of Central Florida, United States\\
{\tt\small Dongdong.Wang@ucf.edu}
\and
Armstrong Aboah\\
Department of Civil, Construction and Environmental Engineering\\
North Dakota State University, United States\\
{\tt\small Armstrong.aboah@ndsu.edu}
\and
Mohamed Abdel-Aty\\
Smart and Safe Transportation (SST) Lab\\
Department of Civil, Environmental and Construction Engineering\\
Joint Appointment Department of Computer Science\\
University of Central Florida, United States\\
{\tt\small M.Aty@ucf.edu}
}

\begin{document}
\maketitle

\input{sec/0_abstract}    
\input{sec/1_intro}

\input{sec/2_related}    
\input{sec/3_method}
\input{sec/4_experiment}
\input{sec/5_conclusion}
{
    \small
    \bibliographystyle{ieeenat_fullname}
    \bibliography{main}
}


\end{document}

%% file: preamble.tex
\usepackage[dvipsnames]{xcolor}


%% file: sec/0_abstract.tex
\begin{abstract}
This paper introduces our solution for Track 2 in AI City Challenge 2024. The task aims to solve traffic safety description and analysis with the dataset of Woven Traffic Safety (WTS), a real-world Pedestrian-Centric Traffic Video Dataset for Fine-grained Spatial-Temporal Understanding. Our solution mainly focuses on the following points: 
1) To solve dense video captioning, we leverage the framework of dense video captioning with parallel decoding (PDVC) to model visual-language sequences and generate dense caption by chapters for video. 2) Our work leverages CLIP to extract visual features to more efficiently perform cross-modality training between visual and textual representations. 3) We conduct domain-specific model adaptation to mitigate domain shift problem that poses recognition challenge in video understanding. 4) Moreover, we leverage BDD-5K captioned videos to conduct knowledge transfer for better understanding WTS videos and more accurate captioning. Our solution has yielded on the test set, achieving 6th place in the competition. The open source code will be available at https://github.com/UCF-SST-Lab/AICity2024CVPRW

\end{abstract}

%% file: sec/1_intro.tex
\section{Introduction}
\label{sec:intro}

Traffic video captioning, an emergent field of video understanding, has received increased attention in recent years due to its capacity to articulate video content through descriptive sentences. This study introduces a dense video-captioning framework, that integrates Parallel Decoding for Video Captioning (PDVC) \cite{wang2021end} with CLIP \cite{radford2021learning} visual features to improve dense captioning of traffic safety videos particularly in scenarios involving pedestrian and vehicle interactions.

Historically, video captioning has been constrained by methodologies that produce overly simplistic and contextually sparse narratives, especially in prolonged and complex traffic videos. Dense Video Captioning (DVC) emerged as a methodological evolution to address these challenges, aiming to provide a comprehensive and detailed narrative of the multifaceted events within traffic scenes \cite{zheng2023chatgpt}, \cite{zheng2023trafficsafetygpt}. Yet, the conventional two-stage approaches to DVC have encountered limitations, primarily due to their segmented process of event localization and subsequent captioning, which often results in a loss of contextual richness and event-specific detail.

In response to this inadequacy, the concept of Parallel Dense Video Captioning (PDVC) has emerged to generate a more coherent narrative by identifying and describing multiple events within a video. PDVC involves two primary tasks: event localization and subsequent event captioning. Traditional approaches to PDVC have adhered to a sequential "localize-then-describe" methodology, initiating with the separation of event boundaries followed by the generation of detailed descriptions. However, this sequential approach has been critiqued for its dependency on the accuracy of event proposal generation, which in turn influences the quality of the captioning. Moreover, the reliance on  anchor designs and post-processing techniques for proposal selection has introduced a plethora of hyper-parameters, complicating the transition to an end-to-end captioning solution.

Addressing these challenges, our research introduces a solution for the AI City Challenge 2024 \cite{Shuo24AIC24} by integrating PDVC with CLIP visual features to improve dense captioning of traffic safety scenario videos, an  end-to-end approach that integrates the localization and captioning processes. PDVC leverages these tasks at the feature level, enabling a more nuanced and accurate event depiction. By employing CLIP to extract frame features, PDVC concurrently decodes these into a set of events with corresponding locations and captions, facilitated by parallel prediction heads. This methodology is enhanced by using an event counter, which refines the prediction of event quantity, thereby mitigating caption redundancy and ensuring a comprehensive video narrative. Therefore, the main contributions of our paper are stated as follows:
\begin{itemize}
  \item Our study introduces a solution by integrating PDVC with CLIP visual features to improve dense captioning of traffic safety scenario videos.
  \item Given domain shift challenge, we conduct domain-specific model adaptation by domain-specific training and knowledge transfer to alleviate domain shift in video context.
  \item This solution examines different impact factors for performance variation.
  \item Experiments show that our proposed solution achieves 6th place on the testing set of the challenge.
\end{itemize}

Our empirical evaluations, conducted the WTS dataset developed by Woven by Toyota, Inc., and BDD-5K dataset, demonstrate PDVC's capacity in Traffic Safety Description and Analysis which underscore the framework's efficacy in generating precise and meaningful video captions, fostering a deeper understanding of the video content. Hence, this study contributes to the field of video captioning by presenting PDVC as a streamlined, effective, and end-to-end solution to addressing real-world challenges encountered in dense traffic video captioning tasks.

%% file: sec/2_related.tex
\section{Related Works}
\subsection{Dense Video Captioning}

Dense video captioning is a complex task involving both event localization and captioning. One  pioneering work by \cite{krishna2017dense} introduced a dense video captioning model featuring a multi-scale proposal module for localization and an attention-based LSTM for context-aware caption generation. Subsequent studies have aimed to improve event representations through context modeling \cite{wang2018bidirectional, yang2018hierarchical}, event-level relationships \cite{wang2020event}, and multi-modal feature fusion \cite{iashin2020better, iashin2020multi}. Further research endeavors focus on enhancing the integration between localization and captioning modules \cite{li2018jointly, zhou2018end}, with proposals like \cite{mun2019streamlined} aiming to boost the efficiency of proposal generation to enhance the coherence of generated captions. An efficient solution to this problem is PDVC\cite{wang2021end}, which parallelizes localization, selection, and captioning tasks within a single end-to-end framework. PDVC simplifies the pipeline while ensuring the generation of accurate and coherent captions. Leveraging the Detection Transformer \cite{carion2020end}, PDVC enhances object detection by attending to sparse spatial locations of images and incorporating multi-scale feature representation. We adopted PDVC as a backbone model to solve dense video caption generation.

\subsection{Domain-specific Learning}

Domain-specific learning, encompassing domain modeling \cite{zhou2022domain} and transfer learning \cite{pan2009survey}, has emerged as a crucial strategy to address the challenge of effectively training models amidst internal data shifts. Often, collected data lacks careful curation, leading to significant variations in features across different segments of the training dataset. This internal shift impedes efficient training and convergence, resulting in degraded model performance on generalized tasks \cite{zhou2022domain, rostami2023overcoming}. To overcome this challenge, one approach involves segmenting the data and deriving domain-specific models \cite{d2019domain}, which helps mitigate domain shift issues. Another approach is transfer learning that facilitates knowledge transfer through pre-training and fine-tuning \cite{pan2009survey}. The effectiveness of transfer learning depends on the correlation between the source and target domains. Higher similarity between domains leads to greater transfer efficiency, resulting in improved model performance. By leveraging domain-specific knowledge and transferable features, these approaches contribute to more effective model training and improved performance across various domains. We leverage both domain modeling and knowledge transfer to address efficient dense video captioning.

\subsection{Traffic Safety Description and Analysis}

Traffic safety video captioning can be classified into two main categories: those addressing traffic scenarios and those focusing on driver behaviors \cite{Lu2023JHPFA},\cite{Yang2023DriverGaze}. For traffic scenarios, traditional analytic frameworks have concentrated on sub-tasks such as class detection \cite{Dai23AIC24} and semantic segmentation \cite{Qin2022IDYOLO}, \cite{inproceedingsShoman}, accident identification \cite{Fang2022Traffic}, risk evaluation \cite{Li2023Anomaly}, lane detection \cite{W2019End}, driver attention monitoring \cite{Aboahetal}, and traffic flow analysis \cite{shoman3},\cite{articleShoman}. While some methodologies are capable of annotating or captioning traffic scenes within static images \cite{Li2022Caption}, \cite{Song2021Two}, they fall short in capturing the dynamic evolution of traffic scenarios over time, lacking the ability to detail incidents with spatial-temporal attributes. Furthermore, there exist methodologies targeting distinct aspects, like polarization feature-based perception \cite{Huifeng2022Vehicle} and action recognition \cite{Xu2022Action},\cite{inproceedingsShoman2}, yet these are not directly applicable for spatial-temporal scenario comprehension.

Recent advancements in video captioning are categorized into template-based and neural network-based methods. Template-based captioning \cite{Xu2015Jointly}, \cite{Guadarrama2013YouTube}, an earlier approach, employs manually defined templates into which video elements are classified and detected to generate descriptive sentences. Despite their efficiency with smaller datasets, these methods often produce rigid and inflexible sentences. In contrast, neural network-based captioning \cite{venugopalan2015translating} benefits from deep learning's progression, enabling complex spatio-temporal feature extraction from both images and videos, and translating these directly into narrative text. The role of attention mechanisms in video captioning is crucial, balancing the local and global features within videos to identify the most relevant temporal segments for narration. Innovations in attention mechanisms have not only improved traditional applications but also shown promise in addressing traffic scenario-specific challenges \cite{Wang2018Deep}, \cite{Wang2020Paying}, \cite{Tan2023VMLH}, \cite{J2023Graph}.

Our method differentiates itself by distinctly addressing different elements in the traffic environment where, separate training is conducted for vehicle and pedestrian models within each domain-specific framework. To ensure consistent caption generation over time, video synchronization is achieved through video trimming. Furthermore, a knowledge transfer process is implemented, significantly improving the video understanding and caption generation capabilities. These models are then applied to generate captions for videos, with a subsequent post-processing phase aimed at enhancing the textual fluency of the generated captions.

%% file: sec/3_method.tex
\section{Methodology}

\textbf{Problem Formulation.} In this task, we explore the solution $s(v_i, c_i)$, aiming to minimize semantic disparities between generated captions ($gc_i$) and actual captions ($c_i$) corresponding to event clips ($v_i$) and text captions ($c_i$) respectively, within a video dataset ($V$). The optimization process involves maximizing the probability $P(c_i | v_i)$ while ensuring temporal alignment, guaranteeing that each caption is appropriately synchronized with its corresponding event window. Notably, the probability $P(c_i | v_i)$ is determined through a generative process, where each word ($w_n$) in the caption sentence ($c_i$) is inferred sequentially. This inference follows a sequential probability chain: $P(c_i | v_i) = P(w_n|w_{n-1}, w_{n-2}, ... v_i)P(w_{n-1}|w_{n-2}, ... v_i) ...P(w_1|v_i)$. This approach enables the systematic examination of caption generation in video contexts, ensuring coherence and alignment between generated and ground truth captions.

\begin{figure*}[htbp]
    \centering
    \includegraphics[width=1\textwidth]{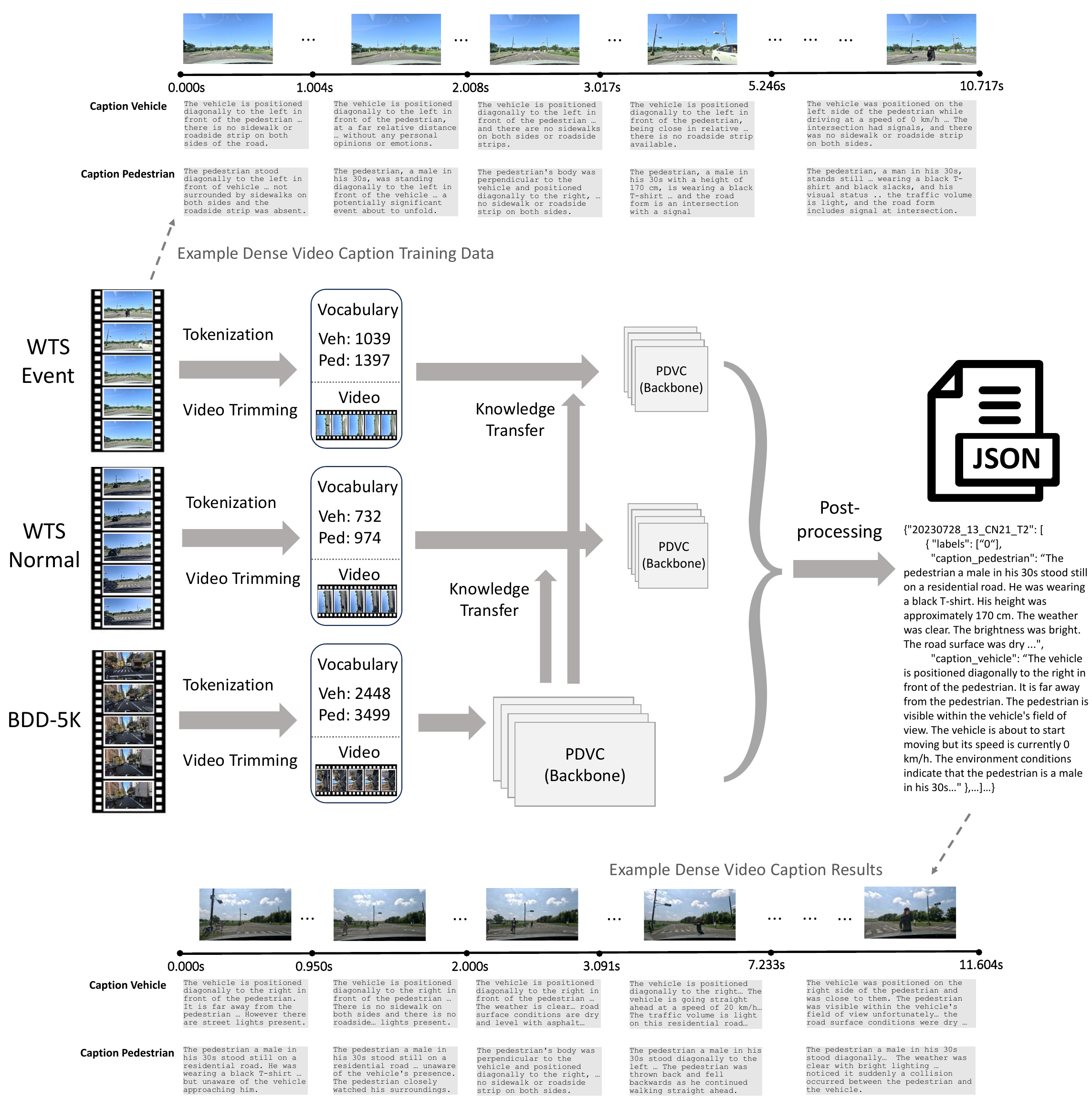} 
    \caption{Overview of our proposed solution. Domain modeling is performed with domain-specific tokenizations for WTS-event, WTS-Normal, and BDD-5K. Caption vehicle (Veh) and caption pedestrian (Ped) models are separately trained with each domain-specific model. To facilitate the caption generation consistency over time, video synchronization is conducted by video trimming. Knowledge transfer from the BDD-5K model to WTS data modeling is conducted to enhance video understanding and caption generation. Subsequently, the models are utilized to infer captions for videos, followed by post-processing that enhances text fluency.}
    \label{fig:overiew}
\end{figure*}

\subsection{Overview}

This section introduces our proposed solution for dense video captioning in traffic safety scenario analysis, comprising four core components. These include data pre-processing, image-text feature preparation, domain-specific model adaptation, and context post-processing. In domain-specific model adaptation, we utilize domain-specific training and knowledge transfer methods to mitigate domain shift across video contexts. By integrating these components, our solution aims to effectively generate descriptive captions that accurately depict traffic safety scenarios.

\subsection{Pre-processing}

All raw data undergo a thorough sanity check and alignment between images and context. We utilize tokenization to prepare text features and employ video trimming to ensure multi-camera synchronization. 

\textbf{Tokenization.} Utilizing all textual descriptions, we constructed a vocabulary comprising a collection of tokens tailored for streamlined tokenization processes. Through experimentation, we observed that vocabulary pruning significantly improves the efficiency of context comprehension and model training. By selectively removing less informative or redundant tokens from the vocabulary, we optimize the tokenization process, thereby enhancing the overall performance of the model. This refined vocabulary enables more effective representation learning, facilitating better understanding of contextual nuances and improving the model's ability to capture relevant information from textual inputs. The pruning process utilizes domain-specific caption datasets. For instance, the  Woven Traffic Safety (WTS) dataset relies on a vocabulary consisting of the unique words from all captions, while BDD-5K utilizes its own dictionary for this purpose.

\textbf{Video Synchronization.} The problem of misalignment between vehicle-view and overhead-view videos poses a challenge in ensuring accurate representation of events across different camera perspectives. To resolve this issue, we implement a video trimming approach guided by caption json files. By aligning the video content with event timestamps specified in the caption files, we synchronize scenarios depicted in the footage captured from various camera angles. This synchronization process ensures consistency in the portrayal of events across different viewpoints, enhancing the overall coherence and accuracy of the video data. Through meticulous alignment of video segments based on caption-guided trimming, we aim to mitigate discrepancies and discrepancies in the depiction of scenes, facilitating modeling on the captured footage across multiple camera views.

\subsection{Feature extraction}

We leverage CLIP \cite{radford2021learning} to extract video features to align with caption text features for video captioning training. The CLIP model represents a state-of-the-art approach for extracting video features, leveraging vision and language representations to understand video content comprehensively. It leverages Convolutional Neural Networks (CNNs) and Transformer-based architectures to encode both visual and textual information effectively. By jointly pre-training on large-scale image-text pairs, CLIP model can more effectively associate visual concepts with their corresponding textual descriptions, enabling it to understand the content of images and videos in a semantically rich manner.

\subsection{PDVC-equipped dense captioning model}

    \textbf{Preliminary: PDVC.} The model of PDVC \cite{wang2021end} stands as a potent solution for generating dense captions in streaming videos. The methodology revolves around crafting a model architecture featuring a sequential image feature encoder and a parallel textual segment decoder. Central to its operation is the integration of the Deformable Transformer \cite{carion2020end}, which facilitates deformable attention mechanisms in both the encoder and decoder modules. The encoder stage capitalizes on a combination of CNNs and Transformer layers to process multi-scale frame features, which are then combined with positional embeddings. These enhanced features undergo multi-scale frame-frame relationship extraction within the deformable transformer encoder. Meanwhile, the decoding network consists of a deformable transformer decoder alongside three parallel heads: a captioning head responsible for generating captions, a localization head dedicated to predicting event boundaries and confidence scores, and an event counter tasked with estimating the appropriate event count. The localization head undertakes box prediction and binary classification for each event query, while the event counter utilizes a max-pooling layer and a fully connected layer with softmax activation to predict the event number. Final outputs are determined by selecting the top N events based on accurate boundaries and favorable caption scores from N event queries. During training, the loss function comprises a weighted combination of segment IOU loss, classification loss, countering loss, and caption loss (as shown in equation \ref{eq:loss}). The overall loss is computed as the sum of training losses across all decoder components. In our inference process, we use the proposals' centers as reference points to automate the final event detection selection. We select the closest segments based on proximity, which are then outputted as captions.

\begin{equation}
    L = \beta_{giou}L_{giou} + \beta_{cls}L_{cls} + \beta_{ec}L_{ec} + \beta_{cap}L_{cap},
    \label{eq:loss}
\end{equation}

where $L_{giou}$ represents the generalized IOU between predicted temporal segments and gournd-truth segments, $L_{cls}$ represents the focal loss between the predicted classification score and the ground-truth label, $L_{ec}$ represents the cross-entropy loss between predicted count distribution and the ground truth, and $L_{cap}$ represents the cross-entropy between the predicted word probability and the ground truth normalized by the caption length.

\subsection{Post-processing}

Post-processing was conducted on the generated text to address units such as km/h, which were not present in the vocabulary. This formatting step aimed to enhance the clarity of the text by reducing the occurrence of unknown tokens, thereby improving the overall understanding and interpretability of the generated content. Through handling of units and minimizing unknown tokens, the processed text gains coherence and semantic clarity, aligning more closely with the intended meaning and enhancing the human-like fluency performance.

%% file: sec/4_experiment.tex
\section{Experiments}
\subsection{Dataset}

The WTS Dataset, developed by Woven by Toyota, Inc., captures detailed behaviors of vehicles and pedestrians in staged traffic events, including accidents. With over 1.2k video events spanning 130 distinct traffic scenarios, WTS integrates multiple perspectives from vehicle ego and fixed overhead cameras. Each event includes comprehensive textual descriptions of observed behaviors and contexts. Additionally, detailed textual description annotations for approximately 5k publicly sourced pedestrian-related traffic videos from BDD100K \cite{yu2020bdd100k}, i.e., BDD-5K are provided for diverse experimental purposes, serving as valuable training and testing resources.

\subsection{Evaluation Metrics}

In this solution, we introduce four metrics—BLEU, ROUGE, METEOR, and CIDEr—for evaluating model performance. These metrics provide comprehensive assessments of the solution's effectiveness across various natural language processing tasks, including machine translation, text summarization, and image captioning.

\textbf{BLEU Score.} The BLEU score (Bilingual Evaluation Understudy) is a metric used to evaluate the quality of machine-translated text by comparing it to one or more reference translations. It measures the similarity between the candidate translation and the reference translations based on n-grams. The BLEU score is calculated using the following formula:
\begin{equation}
    \text{BLEU} = \text{BP} \times \exp\left(\sum_{n=1}^{N} \frac{1}{N} \log p_n\right),
    \label{eq:bleu}
\end{equation}
where $\text{BP}$ is the brevity penalty and $p_n$ is the modified precision for $n$-grams. We adopted BLEU-4 referring to the use of 4-grams, that indicates the metric evaluates the precision of the generated translation in terms of matching 4-word sequences with those found in the reference translations.

\textbf{ROUGE Score.} The ROUGE score (Recall-Oriented Understudy for Gisting Evaluation) is a set of metrics used to evaluate the quality of text summaries by comparing them to reference summaries. It measures the overlap between the candidate summary and the reference summaries in terms of n-grams, word sequences, and sentence-level structures. We employ ROUGE-L, where L stands for "Longest Common Subsequence", to assess the quality of the summary by considering the longest sequence of words that appear in both the generated summary and the reference summaries.

\textbf{METEOR Score.} METEOR (Metric for Evaluation of Translation with Explicit Ordering) is a metric used to evaluate machine translation output by considering both the precision and recall of matching words between the candidate translation and reference translations. The METEOR score is calculated using the following formula:
\begin{equation}
   \text{METEOR} = \frac{\text{Precision} \times \text{Recall}}{\text{Precision} + \alpha \times \text{Recall} + (1 - \alpha)}
   \label{eq:meteor}
\end{equation}

\textbf{CIDEr Score.} CIDEr (Consensus-based Image Description Evaluation) is a metric specifically designed for evaluating the quality of image descriptions generated by image captioning models. It measures the consensus between the generated descriptions and human reference descriptions using n-grams and term frequency-inverse document frequency (TF-IDF) weighting.

\subsection{Implementation Details}

\textbf{General Configuration.} All the training images are resized to 224$\times$224 and normalized. In PDVC, the deformable transformer uses a hidden size of 512 in MSDAtt layers and 2048 in feed-forward layers. The number of event queries is 10. The caption generation module is the vanilla LSTM captioner where hidden dimension in captioning heads is 512. For the event counter, we choose the maximum count as 10. The generated caption length is set to 200 words. We use Adam as the optimizer with initial learning rate set to 1e-3. We train the model with 30 epochs with a batch size of 4 for WTS and 8 for BDD-5K . All models are trained on one GPU Titan RTX.

\textbf{Domain Modeling.} 
In domain modeling, we partition the WTS Normal and WTS Event training sets to develop distinct domain-specific language models.We tokenize the provided captions accordingly. For the caption vehicle, Normal captions require 732 words, while Event captions need 1039 words. For the caption pedestrian, the difference is more significant, with Event captions containing 1397 words compared to 974 words for Normal captions. This variance underscores the significance of domain modeling across these scenarios. Moreover, the video feature difference will further increase this disparity. For BDD-5K, all captions are utilized for model training, with 2448 words required for the caption vehicle and 3499 words for caption pedestrian.

\textbf{Knowledge Transfer.} To harness the superior features of BDD-5k due to its larger dataset, we employ knowledge transfer to enhance the domain models of WTS. We leverage model pretraining with BDD-5K and fine-tune the pretrained model with WTS Normal and Event datasets. To ensure effective transfer, we align the vocabulary sets of BDD-5K and WTS. During the fine-tuning stage, we decrease the learning rate for WTS Normal and Event to 5e-4 for better tuning results. Other training settings follow general configuration.

\subsection{Challenge Results}

\begin{figure*}[htbp]
    \centering
    \includegraphics[width=1\textwidth]{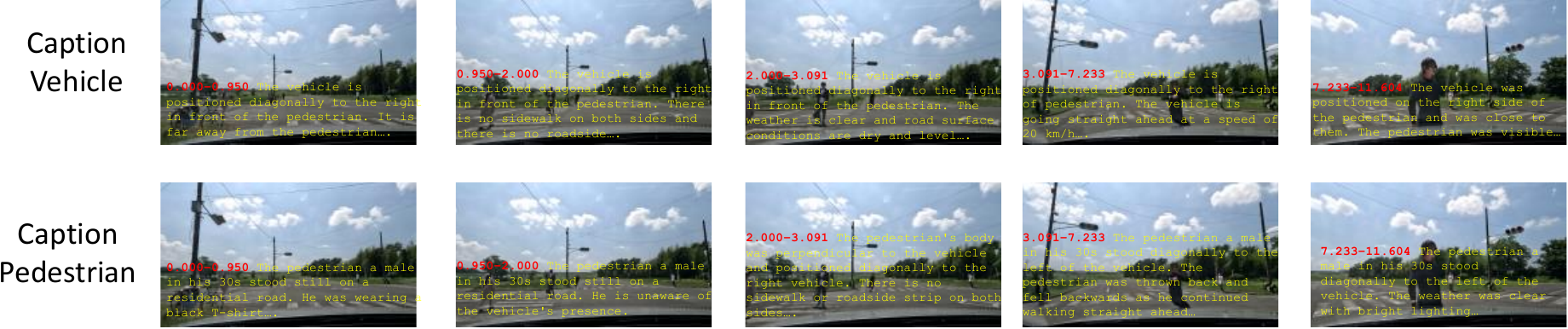} 
    \caption{A video captioned with the proposed solution. The video is divided into five segments based on event and time, and the best proposal captions are matched to each segment's time frame. Due to space limit, only first two sentences are shown. The time is segmented by seconds.}
    \label{fig:video}
\end{figure*}

We conduct video captioning separately for pedestrian and vehicle captions. The BDD-5K  model is trained exclusively on BDD-5K  datasets, whereas the WTS model is initially pretrained using BDD-5K  datasets and then fine-tuned using WTS datasets. The top-performing model achieves an average performance score of 29.0084, with contributions of 30.2821 from external data of BDD-5K  and 27.7347 from internal data of WTS. Additional details regarding the performance breakdown are provided in the Table \ref{tab:performance}. As shown in Table \ref{tab:leaderboard}, the final score of our proposed solution (Team ID 219) on the whole test set is 29.0084.
Our team achieved rank 6 on Track 2 Traffic Safety Description and Analysis of AI City Challenge 2024. We also illustrate one example video presentation result in Figure \ref{fig:video}

\begin{table}[htbp]

    \centering
    \setlength{\tabcolsep}{1.5pt}
    \begin{tabular}{c|ccccc}
    \hline
Data & BLEU-4 & METEOR & ROUGE-L & CIDEr & S2 \\
\hline
 WTS & 0.2005 & 0.4115	&0.4416&	0.5573&	27.7347 \\
 BDD-5K  & 0.2102 &	0.4435	& 0.4705	&0.8698&	30.2821	\\
        \hline
    \end{tabular}
    \caption{Performance on trafffic safety description and analysis task.}
    \label{tab:performance}
\end{table}

\begin{table}[htbp]

    \centering
    \setlength{\tabcolsep}{1.5pt}
    \begin{tabular}{c|ccc}
    \hline
Rank & Team ID & Team Name & Score \\\hline
1 &	208 & AliOpenTrek &	33.4308 \\
2 & 28 & AIO-ISC & 32.8877 \\
3 & 68	& Lighthouse	& 32.3006 \\
4	& 87	& VAI	& 32.2778 \\
5	& 184	& Santa Claude	& 29.7838 \\
\textbf{6}	& \textbf{219}	& \textbf{UCF-SST-NLP}	& \textbf{29.0084} \\
.. & .. & .. & ..\\\hline
    \end{tabular}
    \caption{Leaderboard of Traffic Safety Description and Analysis.}
    \label{tab:leaderboard}
\end{table}

\subsection{ Ablation Study}

We performed an ablation study to assess the impact of various components on performance variability. Specifically, we evaluated the effectiveness of CLIP feature extraction, domain-specific modeling, knowledge transfer, and post-processing techniques. The results of this analysis are detailed in Table \ref{tab:ablation_ped} and \ref{tab:ablation_veh}, showcasing the influence of each component on overall performance. Our thorough examination revealed that domain modeling and knowledge transfer are the primary drivers of performance improvement, with post-processing providing a moderate boost to all scores. There is a trade-off between CIDEr and the other three scores. Notably, we observed greater improvement in captioning vehicle scenarios (Table \ref{tab:ablation_veh}) compared to captioning pedestrian scenarios (Table \ref{tab:ablation_ped}), particularly when incorporating knowledge transfer. This can be attributed to the similarities in camera views between the BDD-5K and WTS vehicle-view datasets, which both prioritize descriptions of vehicular scenarios. 

\begin{table}[h]
    \centering
    \footnotesize
    \setlength{\tabcolsep}{1.25pt}
    \begin{tabular}{cccc|cccc}
        \hline
       CLIP  & Domain  & Knowledge & Post & BLEU& MET & ROUGE & CIDE \\
       feature & modeling & transfer &-processing & -4 &  -ERO & -L & -r\\
        \hline
        \checkmark & \checkmark & \checkmark & \checkmark &  0.1989 & 0.2701 & 0.2835 & 0.1770 \\
        \checkmark & \checkmark & \checkmark &  & 0.1960 & 0.2667 & 0.2793 & 0.1864  \\
        \checkmark & \checkmark &  &  & 0.1161 & 0.2014 & 0.2247 & 0.1193\\
        \checkmark &  &  &  &  0.1090 & 0.1506 & 0.1749 & 0.1005\\
        \hline
    \end{tabular}
    \caption{Ablation study on caption pedestrian of WTS validation set on the proposed method.}
    \label{tab:ablation_ped}
\end{table}

\begin{table}[h]
    \centering
    \footnotesize
    \setlength{\tabcolsep}{1.25pt}
    \begin{tabular}{cccc|cccc}
        \hline
       CLIP  & Domain  & Knowledge & Post & BLEU& MET & ROUGE & CIDE \\
       feature & modeling & transfer &-processing & -4 &  -ERO & -L & -r\\
        \hline
        \checkmark & \checkmark & \checkmark & \checkmark &  0.3189 & 0.4731 & 0.3221 & 0.3342\\
        \checkmark & \checkmark & \checkmark &  & 0.3146 & 0.4681 & 0.3172 & 0.3423 \\
        \checkmark & \checkmark &  &  & 0.1876 & 0.2261 & 0.2397 & 0.2220\\
        \checkmark &  &  &  & 0.1369 & 0.1867 & 0.2292 & 0.1310\\
        \hline
    \end{tabular}
    \caption{Ablation study on caption vehicle of WTS validation set on the proposed method.}
    \label{tab:ablation_veh}
\end{table}

\subsection{Significance of Model Configuration }

We explored the influence of various configurations on performance outcomes. Our investigation revealed that different depths of transformer decoder layers have varying effects on WTS training performance. Optimal performance was attained with a layer depth of 4, with negligible performance gains observed beyond this point. Similarly, for BDD-5K, the best performance was achieved when the layer depth was 6, with minimal performance gains beyond this point. Setting the LSTM layer to a single layer yielded the best results, as multiple layers led to repetitive sentence generation due to training inefficiency. Regarding batch size, WTS training achieved optimal results with a batch size of 4, whereas BDD-5K performed best with a batch size of 8. This observation may be attributed to the differences in data size and distribution, with larger batch sizes potentially benefiting datasets with larger data sizes such as BDD-5K.

%% file: sec/5_conclusion.tex
\section{Conclusions}

In summary, this paper introduces a solution tailored for Track 2 of the AI City Challenge 2024, focusing on Traffic Safety Description and Analysis. The solution tackles various challenges, including efficient tokenization, CLIP-based feature extraction, domain-specific model training to mitigate domain shift, and the implementation of a post-processing technique to enhance human-like fluency. With these strategies, the proposed framework achieved a commendable performance score of 29.0084 on the test set, achieving the 6th position in the competition. The success of this solution underscores its efficacy in effectively addressing real-world challenges encountered in dense video captioning tasks.